\documentclass[10pt,conference,compsocconf]{IEEEtran}

\usepackage{hyperref}
\usepackage{graphicx}	
\usepackage{orcidlink}

\begin{document}
\title{Sketch2CAD: 3D CAD Model Reconstruction from 2D Sketch \\ using Visual Transformer}
\author{
  Hong-Bin Yang\orcidlink{0000-0001-8422-5263} \\
  École polytechnique fédérale de Lausanne (EPFL) \\
  hong-bin.yang@epfl.ch
}

\maketitle

\begin{abstract}
Current 3D reconstruction methods typically generate outputs in the form of voxels, point clouds, or meshes. However, each of these formats has inherent limitations, such as rough surfaces and distorted structures. Additionally, these data types are not ideal for further manual editing and post-processing. In this paper, we present a novel 3D reconstruction method designed to overcome these disadvantages by reconstructing CAD-compatible models. We trained a visual transformer to predict a "scene descriptor" from a single 2D wire-frame image. This descriptor includes essential information, such as object types and parameters like position, rotation, and size. Using the predicted parameters, a 3D scene can be reconstructed with 3D modeling software that has programmable interfaces, such as Rhino Grasshopper, to build highly editable 3D models in the form of B-rep. To evaluate our proposed model, we created two datasets: one consisting of simple scenes and another with more complex scenes. The test results indicate the model's capability to accurately reconstruct simple scenes while highlighting its difficulties with more complex ones.

\end{abstract}

\section{Introduction}
In the field of architectural design, it is a common practice for architects to brainstorm different possibilities and communicate their design ideas through 2D sketches. These sketches serve as an initial step toward the development of the final design. Once a design direction has been chosen, architects then transfer the intermediate or final decision into a 3D model, which is more visually representative and provides a more detailed understanding of the design.

\begin{figure}[h!]
  \centering
  \includegraphics[width=1\columnwidth]{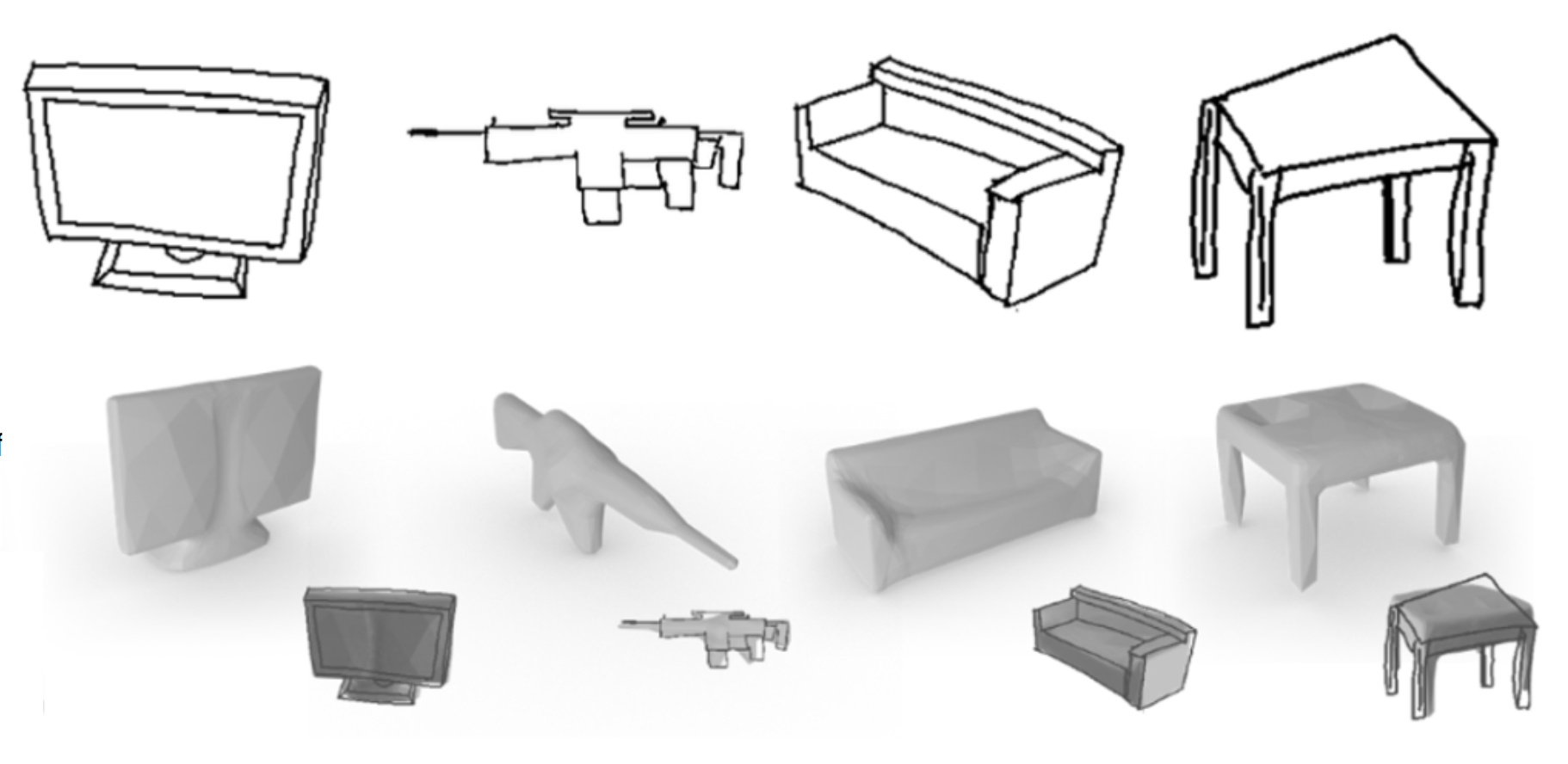}
  \caption{An example of the uneven surface created by the 3D reconstruction method that is based on template-mesh deformation (sketch2model\cite{Zhang_2021_CVPR}).}
  \label{fig:uneven}
\end{figure}

\begin{figure}[h]
  \centering
  \includegraphics[width=1\columnwidth]{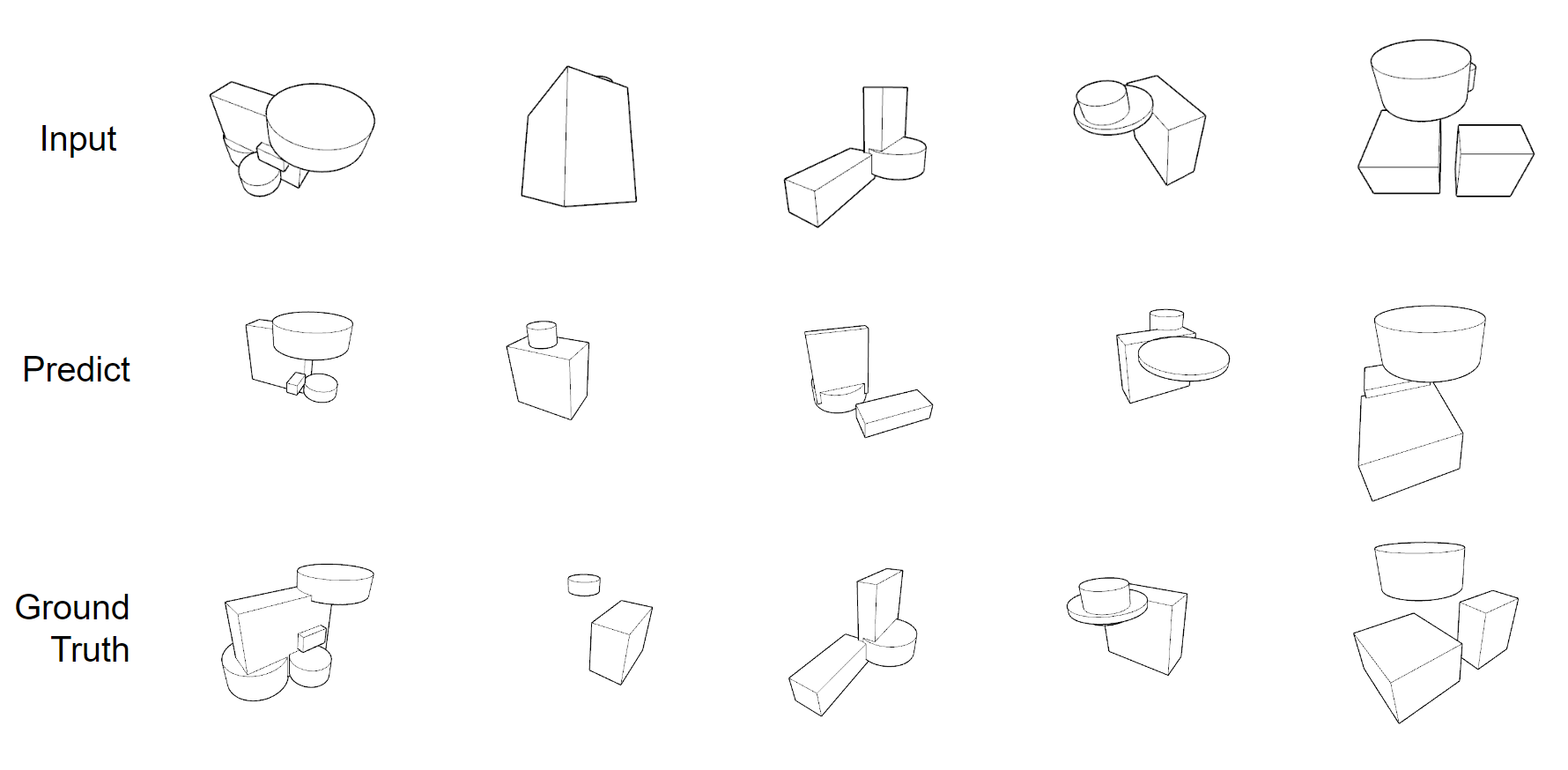}
  \caption{The first row of the example results showcases the input, which is a single 2D wire-frame image. The second and third rows depict the rendered wire-frames from the predicted and ground truth 3D models in the same camera orientation, with the 3D models presented in B-rep form.}
  \label{fig:result_simple_normal}
\end{figure}

Although architects commonly transfer their 2D sketches into 3D models for better visualization and detail, this can be a time-consuming task. While some research has been done on 3D reconstruction from 2D sketches \cite{wang20203d,bhardwaj2022singlesketch2mesh,Zhang_2021_CVPR}, these approaches are not suitable for architectural design. This is because the structures of architecture are often combinations of simple geometric shapes like rectangular boxes or pyramids, which cannot be accurately represented using voxels and point-cloud. While mesh formats may work better, the method based on deforming a template shape, as described in \cite{Zhang_2021_CVPR}, is more effective for objects with curvature and can result in undesired artifacts, such as uneven surfaces and blurred edges, as shown in Figure \ref{fig:uneven}.

Furthermore, the 3D model must also be capable of modification over time as the design is adapted and refined. However, voxel, point cloud, and distorted mesh are not intuitive for humans to interact with and perform manual post-processing. An ideal solution would be to integrate the 3D reconstruction process into 3D modeling software, allowing the generated model to be easily edited. This would also make it easier for the sketch to 3D conversion tool to be utilized effectively within the conventional design pipeline.

\begin{figure*}[t]
  \centering
  \includegraphics[width=2\columnwidth]{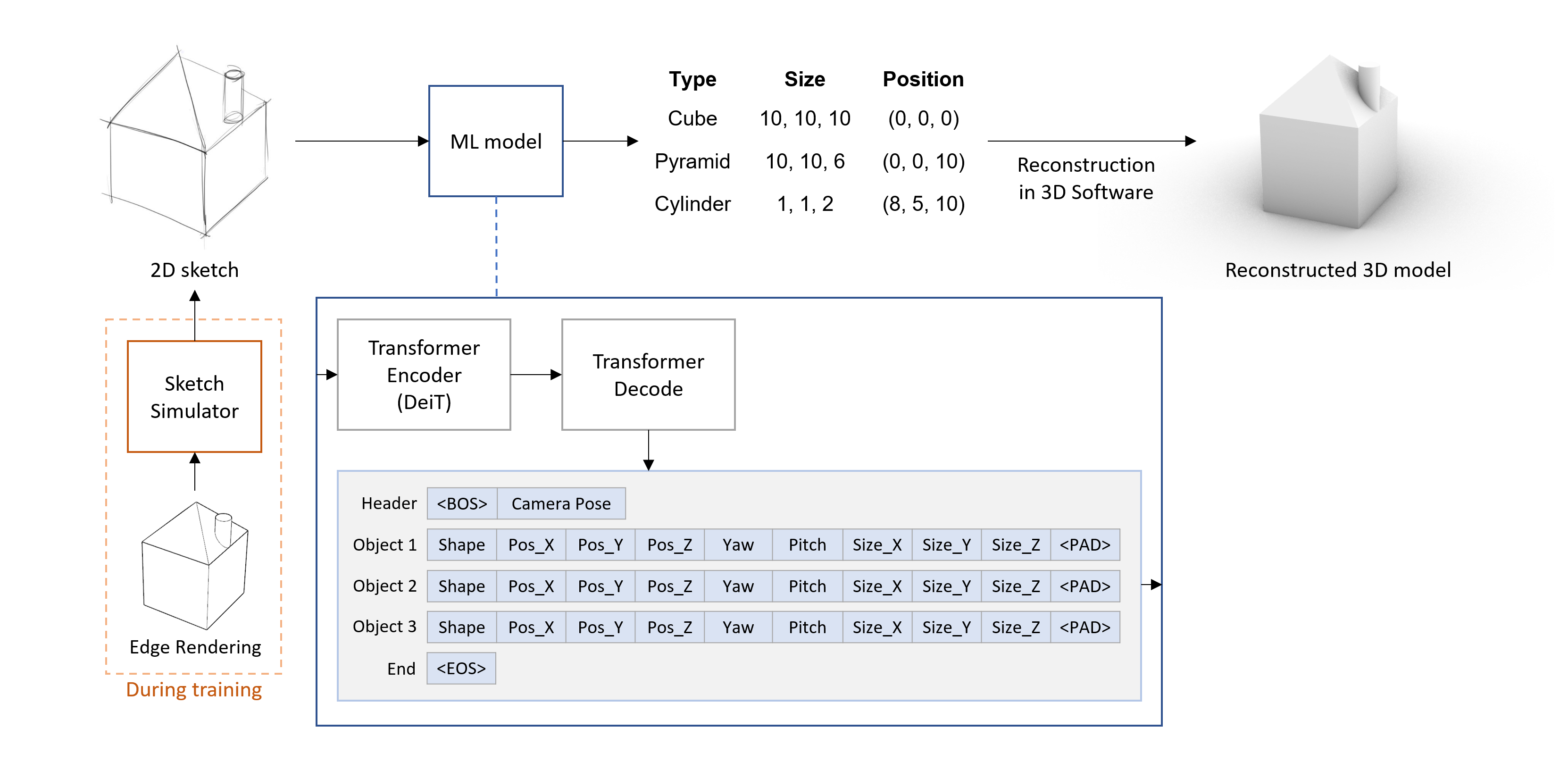}
  \caption{The pipeline of the proposed single image 3D model reconstruction.}
  \label{fig:main}
\end{figure*}

To conclude, the objective of this project is to develop a machine learning model capable of generating a 3D model of architecture from a single 2D hand-drawn sketch, where the result can be seamlessly integrated into conventional 3D modeling software. To achieve this, an visual transformer is trained to take an image as input and generate a sequence of "scene descriptors" containing a list of all the objects appearing in the scene, along with their shape corresponding parameters like position, orientation, and size. To convert the predicted parameter into 3D objects, we programmed Rhino Grasshopper, a widely-used 3D modeling software in architecture design, to read the output and construct the scene accordingly.  Figure \ref{fig:main} shows the overall pipeline of the proposed approach.

Although this project aims to accelerate the 3D modeling process and improve the overall user experience in architectural design, however, generating simplified scene descriptors can also benefit robots that rely on computer vision. If a rough 3D scene can be reconstructed from an RGB image, robots can better navigate themselves and interact with different objects, increasing efficiency and accuracy in various tasks.

\section{Related Work}
The proposed project, despite its primary focus on 3D reconstruction, can be viewed as a fusion of semantic segmentation, object classification, and 6DoF estimation using a distinctive approach. It is worth noting that ML-based 3D reconstruction is an active area of research, and various techniques and methodologies have been proposed to tackle the challenges associated with it.

\subsection{3D Reconstruction}
Training an end-to-end machine learning model is the most common approach for 3D reconstruction today, especially with the presence of large-scale dataset\cite{chang2015shapenet, Sun_2018_CVPR} and differential rendering\cite{Kato_2018_CVPR, liu2019softras}. These learning-based methods are springing up today\cite{han2019image}; some take a single image as input, and some require multiple photos from different views. The model can be generated as voxel\cite{choy20163d,knyaz2018image}, polygon mesh\cite{wang2018pixel2mesh}, and point cloud\cite{ping2022visual,fan2017point,gadelha2018multiresolution}.

However, such ML-based 3D reconstruction methods are usually hampered by low generalization ability, which is an intrinsic issue related to the machine learning model. Since the overall structure of the generated models is limited, these methods typically train class-specific models for objects within the same category. As a result, they can only generate models within the same class with similar structures. For example, if a photo of a cat is fed into a model that is trained to generate cars, it will still generate a car regardless of how different they look. To tackle this problem, \cite{thai20213d,zhang2018learning} suggest using a 2-stage process by estimating the normal map, depth map, and silhouette as intermediate results through the first network, using them as the input to the 2nd network to generate the final 3D model.

This project, in contrast, adopts an entirely different approach to reconstructing 3D objects and thus has the potential to reconstruct a wider range of contexts.

The most related work to the proposed project is Sketch2CAD\cite{li2020sketch2cad}, which allows the user to draw the object's wireframe, and the system will automatically translate it to CAD operations. However, this means that the user has to draw very precisely, including all the hidden lines, making it different from the initial goal of this project, where the expected input is a hand-drawn sketch.

\subsection{6DoF Estimation}
6DoF estimation is the process of determining the position and orientation of an object or device in six degrees of freedom (x, y, z, yaw, pitch, roll). However, existing 6DoF estimation algorithms focus on tracing an object with a known 3D model\cite{He_2020_CVPR,tremblay2018deep}, which is different from this task, where the 3D structure is undetermined.

\section{Methods}
\subsection{Data Generation}
To train the model, a Rhino Grasshopper program has been developed to generate synthetic data for training and evaluation. The data generation process includes creating the 3D scene and the corresponding 2D edge rendering. To thoroughly test the machine learning model's performance, two datasets are generated, namely the \textit{simple dataset} and the \textit{complex dataset}\cite{hong_bin_yang_2023_8002232}.

\subsubsection{3D Scene}
\begin{figure}[h]
  \centering
  \includegraphics[width=1\columnwidth]{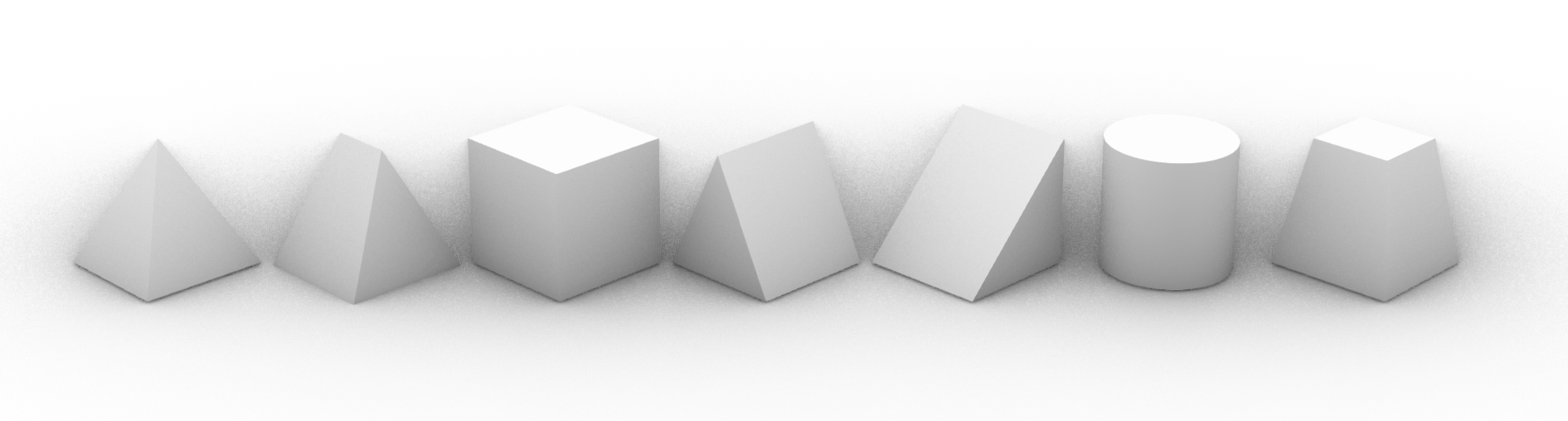}
  \caption{All shapes that appear in the dataset. From left to right are: Pyramid, Hip, Cube, A-frame, Shed, Cylinder, and Mansard.}
  \label{fig:all_shape}
\end{figure}

The 3D scene consists of multiple objects, where the shape type, position, and size are randomly assigned. Since this project aims to improve the architectural design process, the available shapes are chosen based on the appearance of typical residential buildings. Through architectural form analysis, the following shapes were chosen: Cube, Cylinder, Pyramid, Shed, Hip, A-Frame, and Mansard, as illustrated in Figure \ref{fig:all_shape}.

\begin{figure}[h]
  \centering
  \includegraphics[width=1\columnwidth]{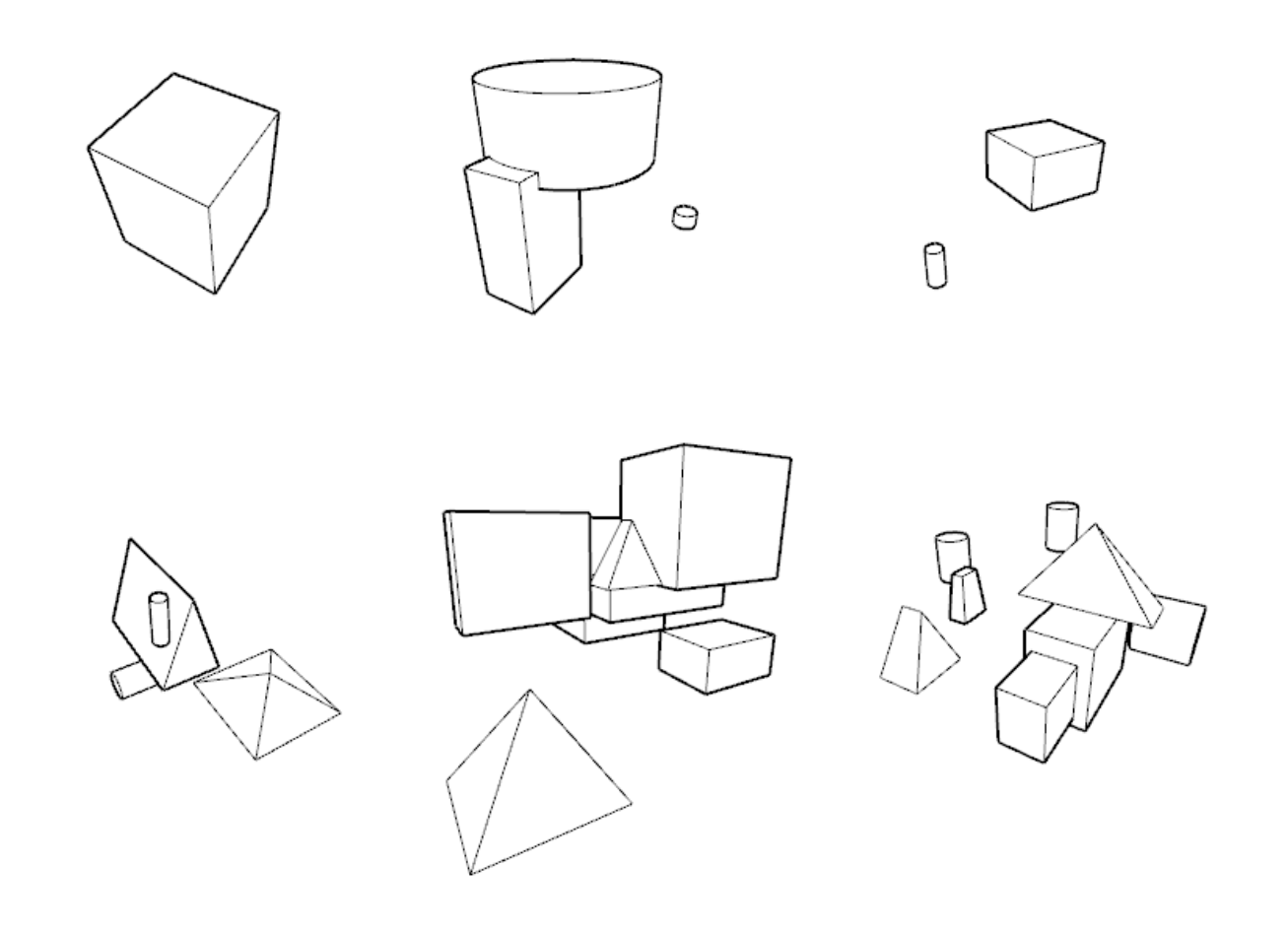}
  \caption{3 examples of the 3D scene from both of the dataset. The first row is the simple dataset, and the second is the complex dataset.}
  \label{fig:dataset_example}
\end{figure}

The \textit{simple dataset} only contains cubes and cylinders. No rotation is performed. Each scene contains 1 to 5 objects. In contrast, the \textit{complex dataset} may have up to 10 objects in all available shapes, with objects randomly rotated at 90°, 180°, and 270° along yaw and pitch. Here, we only use yaw and pitch because of the geometric property of the shape we selected. Figure \ref{fig:dataset_example} showcases examples from both datasets.

To describe the 3D scene, we employ a "scene descriptor" methodology. This descriptor includes the total number of objects in the scene and a comprehensive list of their associated parameters, such as shape, position, rotation angle, and size. These parameters are recorded and exported as a JSON file, enabling seamless data exchange between the machine-learning model and the Grasshopper script.

\subsubsection{2D Sketch}
For each scene, multiple 2D images are rendered from various perspectives. The camera position is determined using the Horizontal Coordinate System, and the perspective is arbitrarily assigned. To maximize diversity and minimize ambiguity, we render 60 images per scene, with varying elevations (ranging from -15° to 45°, every 15°) and azimuths (ranging from -180° to 180°, every 30°).

Two sets of images are rendered to test the precision of the ML model's prediction. The "informative" set contains the shape edge, intersection, hidden lines, and axis highlighting (rendering the x, y, and z axis with red, green, and blue), while the "normal" set contains only the shape edge and intersection. Given the limited time frame, we exclude the informative rendering for the \textit{complex dataset}.

\begin{figure}[h]
  \centering
  \includegraphics[width=1\columnwidth]{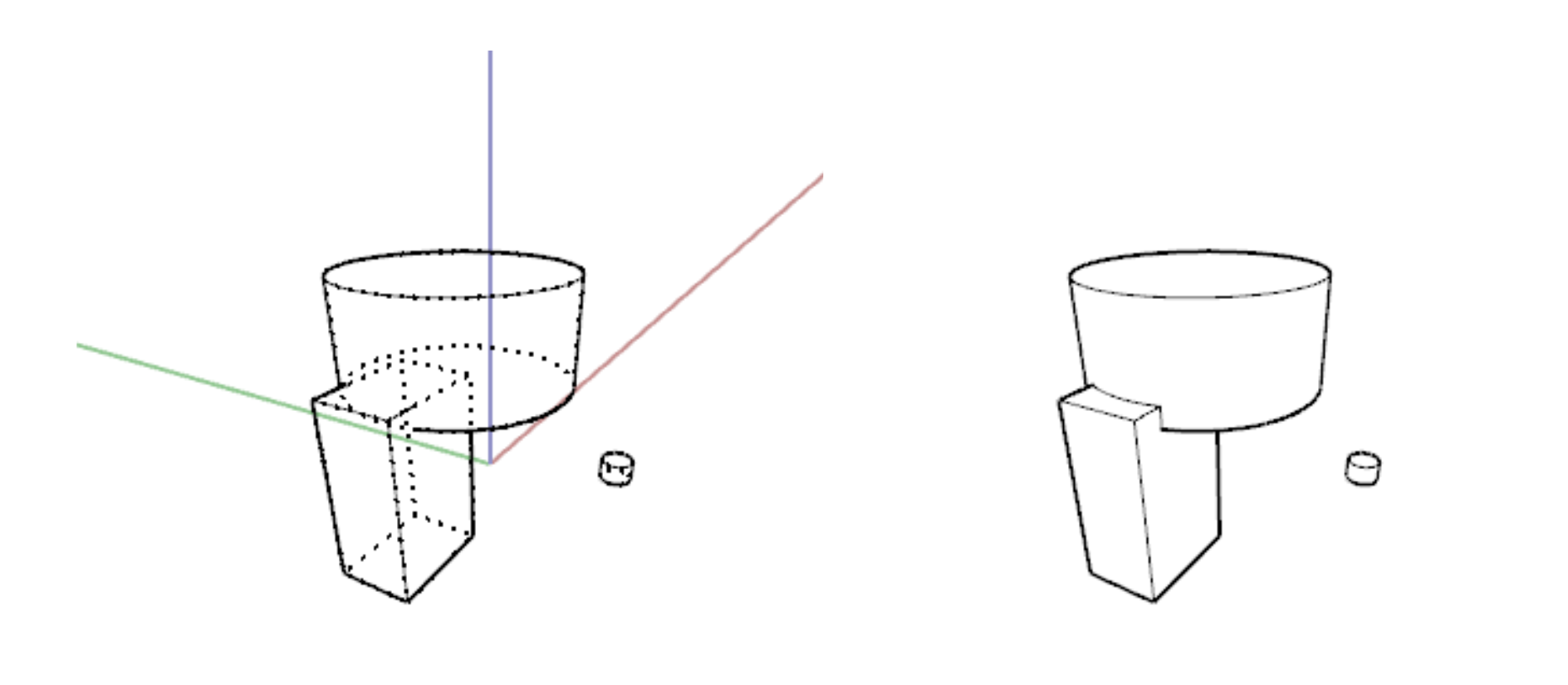}
  \caption{An example of the two type rendering. The left side is the "informative" one, which labels the x-, y-, and z-axis with red, green, and blue, respectively, and the hidden wire-frame is rendered as a dotted line. The right side is the "normal" edge rendering.}
  \label{fig:render_example}
\end{figure}

    



\subsection{Object Classification and Parameter Prediction}
The model is developed based on Pix2Seq\cite{chen2021pix2seq,chen2022unified}, which is a visual transformer \cite{dosovitskiy2020image} framework developed for object detection. It reframes the object detection problem as a text generation task, where the model generates a sequence of tokens describing the objects in an image and their bounding box coordinates. In this project, the program is built on an open-source PyTorch implementation of Pix2Seq \cite{Shariatnia_Pix2Seq-pytorch_2022}.

\subsubsection{Model Architecture}
The model consists of an encoder that reads an image as input and outputs the image embedding and a decoder that generates the final sequence.

For the vision encoder, DeiT-III Small\cite{touvron2022deit} is used, which is designed to be trained with less amount of data, as well as the computational resource and training time. The output is then fed into a vanilla Transformer decoder\cite{vaswani2017attention}, which generates one token at a time, and the next token is predicted based on the preceding tokens and the encoded image representation.

\subsubsection{Sequence construction}
To transform the 3D scene prediction as a text generation task, we must discretely express the parameters in the scene descriptor and assign a corresponding vocabulary as the token.

For the camera position, since the 2D image is rendered from a pre-defined position and angle, we keep a map of $(\mathrm{ID}_{pose}, (\mathrm{azimuth}, \mathrm{elevation}))$ and use the pose ID as the vocabulary directly.

For the object shape, it is naturally a discrete property, so no further conversion is needed. To reconstruct a 3D scene, we need to know each object's position, rotation, and size, which is usually in a continuous domain. To tokenize these values, we adopted a similar strategy as Pix2Seq, by arbitrarily deciding the number of bins for each parameter and uniformly discretizing the value into an integer between $[0, n_{bins} - 1]$. Specifically, the following equation is used for quantization and conversion,
$$
Q_i = \frac{(x_i - \min(X))}{(\max(X) - \min(X))} \times (n_{\mathrm{bins}} - 1)
$$
where the $x$ is the original continuous value, and $Q$ is the quantified embedding. The vocabulary is only shared between different axis in the same property, resulting in the total vocabulary size equal to $n_{\mathrm{cam-pose}} + n_{\mathrm{shape-type}} + n_{\mathrm{bin_{pos}}} + n_{\mathrm{bin_{rot}}} + n_{\mathrm{bin_{size}}}$.

With the above-mentioned conversion method, an object can be represented in the following sequence: [shape-type, position-x, position-y, position-z, yaw, pitch, size-x, size-y, size-z]. To serialize multiple object descriptions and form the scene description, a random ordering strategy is used as Pix2Seq proved that it outperforms other deterministic ordering. The camera pose is encapsulated at the beginning of the sequence, which may benefit the estimation of position, rotation, and size.

\subsubsection{Training Detail}
The objective of the model is to minimize the cross-entropy between the predicted sequence and the groundtruth. During training, the decoder will always see the prior tokens from the groundtruth when predicting the next one.

The encoder's weight is initialized by the weight pretrained on ImageNet-22k and fine-tuned on ImageNet-1k, and the decoder's weight is randomly initialized. We used AdamW optimizer with an initial learning rate equal to $10^{-4}$ and weight decay of $10^{-4}$. Learning rate warmup is used for 15 epochs and then linearly decays the learning rate over the training process. 

To enable the ability to read hand-drawn sketches as input during the inference time, the initial plan is to preprocess the edge rendering with sketch simulator\cite{orbay2014pencil,vinker2023clipascene} before feeding into the network. However, after spending some time trying to run CLIPascene\cite{vinker2023clipascene}, I did not succeed and thus had to postpone this part as future work.

\section{Experiment}
\begin{figure}[h]
  \centering
  \includegraphics[width=1\columnwidth]{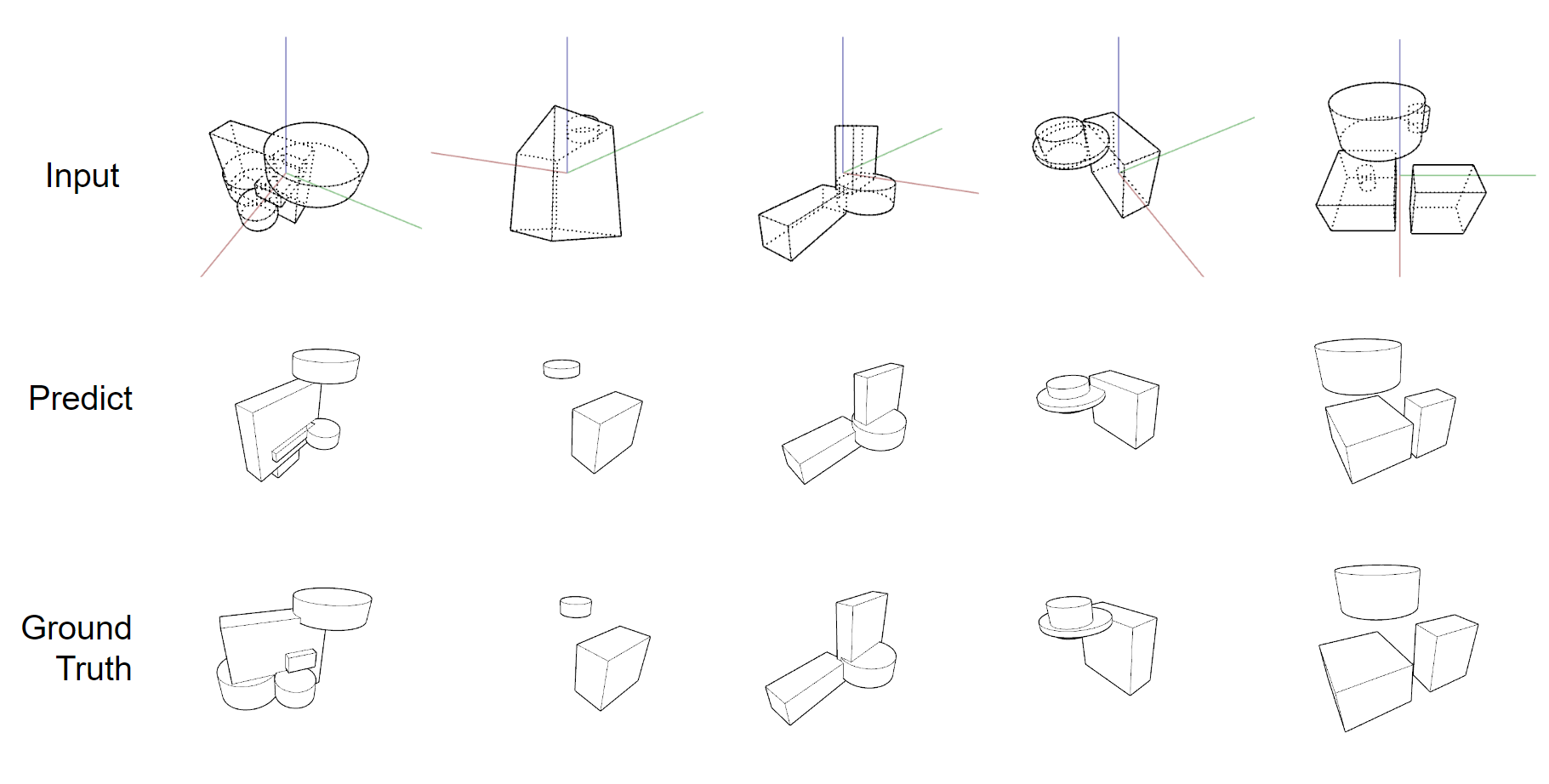}
  \caption{Qualitative result of the simple dataset with informative 2D edge rendering}
  \label{fig:result_simple_informative}
\end{figure}

To begin with, we trained the model with the simple dataset and informative edge rendering. The number of bins for position and size are both set to 20. The loss converges after 105 epochs. From the visual result shown in Figure \ref{fig:result_simple_informative} and the quantitative result in Table \ref{Table:simple}, we can see that the 3D scene is reconstructed precisely. Later, we fine tuned the model with the normal edge rendering, and we noticed a considerable performance drop. Since the model loses the reference to the original point and the axis information, it fails to estimate the camera pose correctly, and the error in position and size estimation is also increased.

\begin{table}[!h]
\begin{center}
\begin{tabular}{ | c | c | c | } 
  \hline
  & Informative & Normal\\
  \hline
  Camera Pose Estimation (Acc) & 0.99 & 0.21 \\ 
  \hline
  Object Classification (F1-score) & 0.98 & 0.93\\ 
  \hline
  Position error (world size = 20) & (2.50, 2.74, 1.65) & (5.94, 4.93, 2.13)\\ 
  \hline
  Size error (max = 20) & (1.07, 1.11, 0.84) & (1.81, 1.75, 1.12)\\
  \hline
\end{tabular}
\caption{\label{Table:simple} Result on simple dataset}
\end{center}
\end{table}

\begin{figure}[h]
  \centering
  \includegraphics[width=1\columnwidth]{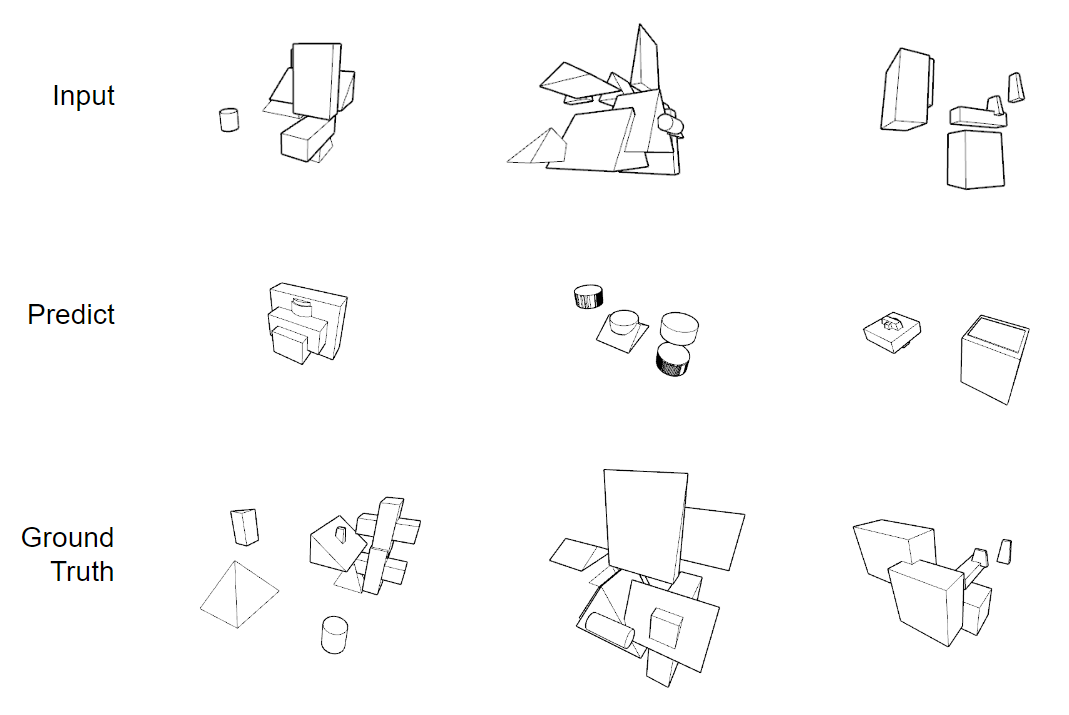}
  \caption{Qualitative result of the complex dataset with normal 2D edge rendering}
  \label{fig:result_complex_normal}
\end{figure}

With the success of training the simple dataset, we moved to the more complex scenes. In this experiment, the number of bins for position is set to 200, for size is 60, and for rotation is 4 (since we only consider 0°, 90°, 180°, and 270°). However, the model fails to generate a satisfactory result. From the visual result shown in Figure \ref{fig:result_complex_normal}, we can see that the predicted scene has nothing to do with the input image, not to mention the overall scene. 

Initially, we suspect that the model actually predicted a scene that matches the perspective as the input image since the scene is reconstructed from a single image. However, it is not the case if we look closely -- the shapes are not even classified correctly. In the center column of the visual result, we can see that the predicted 3D scene mainly consists of cylinders, but there is only one cylinder in the input image and the rest are triangular shape.

For the reason of failing, one guess is that the scene is too complex and beyond the model's capability, or we did provide enough data for it to train. Also, it is easier to find occlusion in the complex dataset since the maximum object size and the number of objects in each scene are increased. Such occlusion may produce too much noise during the training time. Lastly, since the synthetic dataset are generated randomly, the lack of context may also be an issue.

\begin{table}[!h]
\begin{center}
\begin{tabular}{ | c | c | } 
  \hline
   & Normal\\
  \hline
  Camera Pose Estimation (Acc) & 0.5 \\ 
  \hline
  Object Classification (F1-score) & 0.65\\ 
  \hline
  Position error (world size = 200) & (54.35, 62.65, 49.57)\\ 
  \hline
   Rotation error (360°) & (12.86, 2.14)\\
  \hline
  Size error (max = 60) & (14.21, 13.97, 10.05)\\
  \hline
\end{tabular}
\caption{\label{Table:complex} Result on complex dataset}
\end{center}
\end{table}

\section{Conclusion and Limitations}
In this project, we proposed a transformer-based 3D model reconstruction method, which takes a single image as input and generates a sequence of objects' parameters, which can then be used as input for CAD software and reconstruct the 3D scene.

To train and test this model, we created two datasets with 2 types of edge rendering methods and proved their efficiency and accuracy when presented with a simple scene.

Nevertheless, the proposed method has its limitations mainly in two aspects. First, during our experiment, we saw that the model failed to predict the complex 3D scene. Also, since it can only reconstruct objects with known shapes, even if we can post-process the 3D model with boolean operations and create objects with various shapes and topologies, it is still unlikely for the current model to reconstruct shapes with complex curvature.

3D reconstruction from a single image is still an ill-posed problem. The proposed method tried a novel approach to tackle this problem. Also, leveraging the integration with conventional 3D CAD software increases its potential to be deployed in real-world applications.





\bibliographystyle{IEEEtran}
\bibliography{main.bib}

\end{document}